\pdfoutput=1

\documentclass[11pt, dvipsnames]{article}

\usepackage[]{acl}

\usepackage{times}
\usepackage{latexsym}
\usepackage{float}
\usepackage{graphicx}
\usepackage{xcolor}
\usepackage{soul}
\usepackage{colortbl}
\usepackage{color}

\usepackage{tabularx}
\usepackage{dingbat}
\usepackage{pifont}
\usepackage{amsmath}
\usepackage{amssymb}
\usepackage{makecell}
\usepackage{arydshln}
\usepackage{stfloats}
\usepackage{subfig}
\usepackage[T1]{fontenc}
\usepackage[utf8]{inputenc}

\usepackage{CJKutf8}
\usepackage{multirow}
\usepackage{array}
\usepackage[utf8]{inputenc}
\usepackage{microtype}
\usepackage{booktabs}
\usepackage{xcolor}
\usepackage{fontawesome5}
\usepackage{caption}
\usepackage{amsfonts}
\usepackage{inconsolata}
\usepackage{arydshln}
\usepackage{CJKutf8}
\usepackage{fancyvrb}
\usepackage{listings}

\newcommand{\cmark}{{\textbf{\textcolor[rgb]{0.1, 0.5, 0.1}{\ding{51}}}}}
\newcommand{\xmark}{{\textbf{\color{red}{\ding{55}}}}}

\definecolor{myblue}{rgb}{0.82, 0.94, 0.75}

\title{Error Analysis Prompting Enables Human-Like Translation Evaluation in Large Language Models}

\author{Qingyu~Lu$^{\diamondsuit}$,
\ \textbf{
\ Baopu Qiu$^{\flat}$,
\ Liang Ding$^{\sharp}$, 
\ Kanjian Zhang$^{\diamondsuit\spadesuit}$\thanks{~~Corresponding Author.},
\ Tom Kocmi$^{\heartsuit}$,
\ Dacheng Tao$^{\natural}$} \\
\ $^{\diamondsuit}$Southeast University
\ $^{\flat}$Nanjing University
\ $^{\sharp}$The University of Sydney \\
\ $^{\spadesuit}$Southeast University Shenzhen Research Institute
\ $^{\heartsuit}$Microsoft
\ $^{\natural}$Nanyang Technological University\\
\includegraphics[scale=0.15]{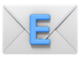} \texttt{luqingyu@seu.edu.cn}, \texttt{qiubaopu@smail.nju.edu.cn}, \\ \texttt{liangding.liam@gmail.com},
\texttt{tomkocmi@microsoft.com}\\
\includegraphics[scale=0.09]{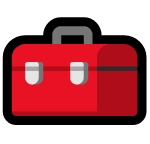} \url{https://github.com/Coldmist-Lu/ErrorAnalysis_Prompt}
}

\begin{document}
\maketitle
\begin{abstract}

Generative large language models (LLMs), e.g., ChatGPT, have demonstrated remarkable proficiency across several NLP tasks, such as machine translation, text summarization. Recent research \citep{kocmi-federmann-2023-large} has shown that utilizing LLMs for assessing the quality of machine translation (MT) achieves state-of-the-art performance at the system level but \textit{performs poorly at the segment level}.
To further improve the performance of LLMs on MT quality assessment, we conduct an investigation into several prompting designs, and propose a new prompting method called \textbf{\texttt{Error Analysis Prompting}} (EAPrompt) by combining Chain-of-Thoughts \cite{wei2022chain} and Error Analysis~\cite{lu-etal-2023-toward}. This technique emulates the commonly accepted human evaluation framework - Multidimensional Quality Metrics (MQM, \citet{freitag-etal-2021-experts}) and \textit{produces explainable and reliable MT evaluations at both the system and segment level}. Experimental Results from WMT22 metrics shared task validate the effectiveness of EAPrompt on various LLMs, with different structures. Further analysis confirms that EAPrompt effectively distinguishes major errors from minor ones, while also sharing a similar distribution of the number of errors with MQM. These findings highlight the potential of EAPrompt as a human-like evaluator prompting technique for MT evaluation. 

\end{abstract}

\section{Introduction}

Large language models (LLMs), especially Generative Pre-trained Transformer (GPT) models~\cite{radford2019language, brown2020language} such as ChatGPT~\citep{ouyang2022training, achiam2023gpt}, have shown remarkable performance in various natural language processing (NLP) tasks~\cite{qin2023chatgpt,zhong2023chat} and complex reasoning and agent applications~\cite{zhong2024achieving,ren2024healthcare}. LLMs are capable of integrating multiple NLP tasks and can generate detailed and comprehensive responses to human inquiries. Additionally, they can respond appropriately to follow-up questions and maintain sensitivity throughout several turns of conversation.

Previous research has demonstrated that LLMs can perform as well as or even better than other LLMs in machine translation task~\cite{hendy2023good,jiao2023chatgpt,Peng2023ChatGPT4MT}. Given the high cost and time-intensive nature of human evaluation, there is a growing demand for MT metrics that offer both explainability and reliability. Therefore, LLMs hold promise in serving as ideal evaluators, capable of generating both judgments and explanations for the translations. 

Concurrent to our research, GEMBA \citep{kocmi-federmann-2023-large} presents an encouraging finding that GPT models can surpass current best MT metrics at the system level quality assessment using straightforward zero-shot standard prompting, confirming the reliability and potential of this technique. However, such prompts exhibit unrealistic performance at the segment level, and cannot offer additional interpretable information regarding translation errors, thus detracting from the goal of achieving a "human-like" evaluation.

\begin{figure*}[t]

\includegraphics[scale=0.35]{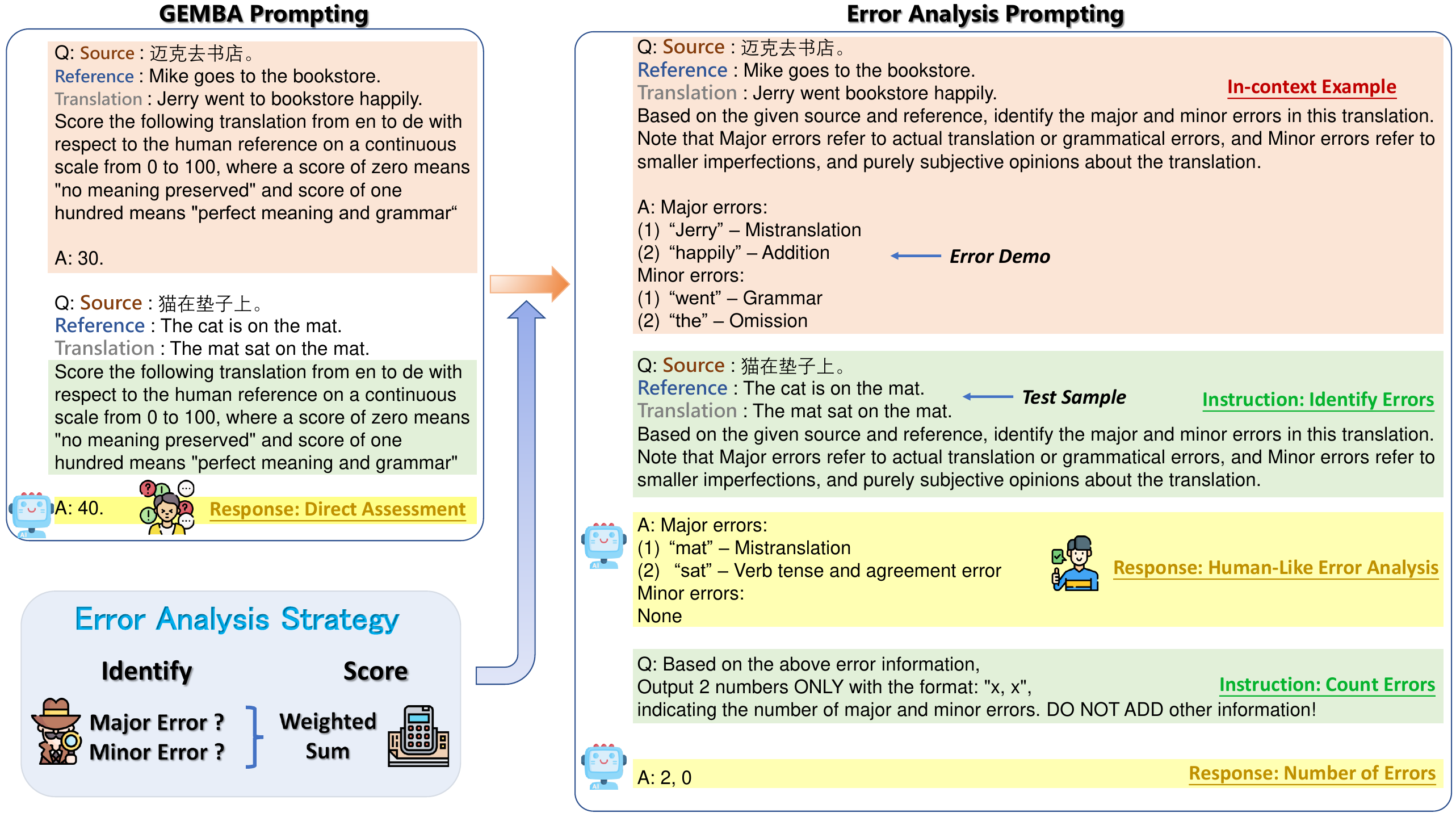}
\centering
\caption{\textbf{A comparative overview between GEMBA Prompting and our proposed Error Analysis Prompting} in assessing the MT quality with LLMs.}
\label{fig:overview}
\end{figure*}

To this end, we take the further step by carefully investigating advanced prompting strategies upon various LLMs for MT quality assessment and propose a novel prompting strategy - \textbf{Error Analysis Prompting} (EAPrompt), combining the Chain-of-Thought (CoT, \citet{wei2022chain}) and Error Analysis (EA, \citet{lu-etal-2023-toward}). 
We give an example of EAPrompt in Figure~\ref{fig:overview}. The idea is to prompt LLMs to emulate the human evaluation framework - MQM \citep{freitag-etal-2021-experts} by \ding{182} \textit{identifying major\&minor errors}, and \ding{183} \textit{scoring the translations according to the severity of these errors}.

We conduct experiments using the test set from the WMT22 metrics shared task, comprising 106,758 segments on 54 MT systems across diverse domains to verify the effectiveness of our approach. Our findings reveal that:

\begin{itemize}
    \item EAPrompt significantly enhances the performance of LLMs at the system level. Notably, prompting \textit{GPT-3.5-Turbo} with EAPrompt outperforms all other metrics and prompting strategies, establishing a new state-of-the-art.
    \item EAPrompt surpasses GEMBA in 8 out of 9 test scenarios across various language models and language pairs, demonstrating superior performance at the segment level.
    \item The findings regarding EAPrompt's strong performance remain consistent even in reference-less settings, highlighting its suitability for quality estimation tasks.
    \item When designing prompts, we recommend the EAPrompt variant featuring a 2-step separated prompting approach and itemized error demonstrations.
    \item Further analysis confirms that EAPrompt adeptly distinguishes major errors from minor ones, closely aligning its error distribution with MQM.
    \item Optimizing the inference costs of EAPrompt can be achieved by leveraging Regular Expressions instead of counting queries. 
\end{itemize}

This study provides an initial exploration of utilizing error analysis to prompt LLMs as evaluators. EAPrompt can also be extended to benefit other evaluation scenarios within language generation, including summarization and data-to-text tasks.

\section{Prompt LLMs with Error Analysis}
\label{sec:evaluation}

\subsection{Translation Evaluation Metric}

Translation evaluation metrics are used to assess the performance of machine translation systems on specific test sets \citep{freitag-etal-2022-results, mathur-etal-2020-results}. These metrics typically take inputs from three sources: the sentence from source language ("Source"), the reference translation provided by human translators ("Reference"), and the hypothesis being evaluated ("Translation"). In scenarios where reference signals are not provided, this "reference-less" metric can also be utilized for quality estimation purposes \citep{zerva-etal-2022-findings, specia2010machine, qiu2022original}. The output of the metric is a score or rank indicating the translation quality of each hypothesis.

To verify the reliability of MT metrics, Multi-dimensional Quality Metric (MQM) has been adopted recently in WMT as a high-quality human evaluation strategy \citep{freitag-etal-2021-experts}. It asks human experts to annotate the errors in the hypothesis and categorize them into "Major" and "Minor" indicating their severity. A detailed description of MQM annotation is presented in Appendix~\ref{appendix:MQM}.

\subsection{Prompt LLMs as Evaluation Metrics}

When prompting LLMs as evaluation metrics, it is crucial to design appropriate instructions that describe the evaluation task. In this paper, we mainly adopt two prompting strategies: "GEMBA Prompting" and "Error Analysis Prompting".

GEMBA \citep{kocmi-federmann-2023-large} is a zero-shot prompting approach that directly asks LLMs to generate a score that reflects the quality of the translation, which shows state-of-the-art performance on GPT models when compared to other model-based metrics. However, they also observe that the performance at the segment level is relatively poorer. This highlights the importance of combining Chain-of-Thought with the Error Analysis Strategy to prompt LLMs in a manner that more closely resembles human evaluation.

\subsection{Error Analysis Prompting}

Motivated by the MQM framework in human evaluation, the idea of the Error Analysis (EA) paradigm, as introduced by \citet{lu-etal-2023-toward}, is to enhance the automatic scoring process by explicitly incorporating error identification, thus providing a more human-like evaluation.

The Chain-of-Thought (CoT) prompting strategy was first proposed by \citet{wei2022chain}. Instead of directly generating the answer, CoT prompts LLMs to think step-by-step. This approach has shown significant performance improvements on reasoning tasks, such as GSM8K \citep{cobbe2021training}. CoT is an emergent ability of LLMs and has been incorporated in instruction fine-tuning of LLMs \citep{chung2022scaling} as well as in benchmarks designed to evaluate LLM capabilities \citep{suzgun2022challenging}.

In this work, we combine the CoT and EA paradigms, introducing a novel prompting strategy called Error Analysis Prompting (EAPrompt). As shown in Figure~\ref{fig:overview}, EAPrompt divides the scoring process into two stages: First, the LLM is instructed to identify major and minor errors in the translation ("Instruction: Identify Errors"). Subsequently, the number of these two types of errors is counted ("Instruction: Count Errors"). Distinguished from GEMBA prompting, EAPrompt emulates the evaluation process of MQM and produces more explainable and reliable automatic evaluations.

After exploring several prompt contexts in initial experiments, we made the following modifications to EAPrompt as follows:

\begin{itemize}
  \item we adopt the one-shot learning format \citep{brown2020language} to enhance the LLMs' understanding of the task (\S\ref{sec:mainres}); different in-context examples are used for different language pairs;
  \item we employ itemized error demonstration in the template response, enabling clearer identification and quantification of errors (\S\ref{sec:ablation});
  \item we partition the evaluation process into two stages to enhance the reliability of metric performance. Additionally, we present a simplified alternative to optimize inference costs by counting errors automatically (\S\ref{sec:repr_exp}).
\end{itemize}

\subsection{Post-processing of LLM responses} \label{sec:postprocess}

After obtaining the number of major and minor errors, we compute the final score of the translation using the following equation:
\begin{equation}
\text{score} = - w_{\text{major}}n_{\text{major}} - w_{\text{minor}}n_{\text{minor}},
\end{equation}
where $n_{\text{major}}$ and $n_{\text{minor}}$ denotes the number of major and minor errors respectively, while $w_{\text{major}}$ and $w_{\text{minor}}$ represent the severity weight assigned to major and minor errors. Since different LLMs may apply distinct criteria for major and minor errors, we follow \citet{lu-etal-2023-toward} to adopt a flexible scoring approach by fixing the $w_{\text{minor}} = 1$ while treating $w_{\text{major}}$ as a latent variable within EAPrompt. We present an analysis on the influence of this variable in \S\ref{sec:major_weight_adjust} and the detailed implementation in experiments is described in Appendix~\ref{sec:appendix_post}.

\begin{table*}[ht]
\centering
\begin{tabularx}{0.94\textwidth}{ccccX}
\toprule[0.5mm]
\textbf{Dataset} & \textbf{Language Pair} & \textbf{Segments} & \textbf{Systems} & \textbf{Domains} \\\midrule
\multirow{3}{*}{WMT22} & En-De & 2037 & 17 & conversational, e-commerce, news, social \\
 & En-Ru & 2037 & 17 & conversational, e-commerce, news, social
\\
 & Zh-En & 1875 & 20 & conversational, e-commerce, news, social \\

\bottomrule[0.5mm]
\end{tabularx}
\caption{\textbf{Statistics of testset}. Source, reference texts, and translations are from the WMT22 metrics shared task. }
\label{tab:experiments}
\end{table*}

\section{Experimental Results}\label{sec:results}

\subsection{Experiment Setup}

\paragraph{Dataset}

We utilize the test set from the WMT22 shared tasks \citep{freitag-etal-2022-results} in English-German (En-De), English-Russian (En-Ru), and Chinese-English (Zh-En) across 4 different domains - conversational, e-commerce, news, and social. This study necessitates the evaluation of 106,758 segments from 54 MT systems. In accordance with GEMBA \cite{kocmi-federmann-2023-gemba}, we deliberately exclude previous datasets such as WMT21 and WMT20. This reduces the potential risk of data contamination, as LLMs may have utilized these datasets in their training sets. Table~\ref{tab:experiments} provides statistics about our test set. For future studies, we plan to incorporate more recent testsets, such as WMT23 \cite{freitag-etal-2023-results}. 

\paragraph{Human Evaluation} 

 We utilize MQM \citep{freitag-etal-2021-experts} as human judgments, which is annotated by human experts and has been widely adopted in recent WMT metrics shared tasks \citep{freitag-etal-2022-results} and quality estimation tasks \citep{zerva-etal-2022-findings}.

\paragraph{Meta Evaluation}

We follow the standard meta-evaluation approach to measure the performance of MT evaluation metrics \citep{freitag-etal-2023-results}. At the system level, we use pairwise accuracy across all three language pairs, which calculates the proportion of all possible pairs of MT systems that are ranked the same by the metric and human scores \citep{kocmi-etal-2021-ship}. At the segment level, we adopt the group-by-item pairwise accuracy with tie calibration as described by \citet{deutsch-etal-2023-ties}. We use the $\text{acc}_{eq}^*$ variant to compare vectors of metric and gold scores for each segment, then average the results over segments. All the meta-evaluation are calculated with MTME\footnote{ \url{https://github.com/google-research/mt-metrics-eval}}, a metric evaluation tool recommended by WMT \citep{freitag-etal-2022-results} to maintain comparability with other metrics.

\paragraph{Significance Analysis}

At the segment level, we follow WMT22 metrics shared task to utilize PERM-BOTH hypothesis test \citep{deutsch-etal-2021-statistical} to assess the significance of metrics. We use 1000 re-sampling runs and set $p=0.05$.

\subsection{Baselines and Large Language Models} 

\paragraph{Baseline Metrics}

Given the reported unreliability of BLEU \citep{papineni-etal-2002-bleu}, we compare our method with several model-based metrics for MT evaluation. BLEURT \citep{sellam-etal-2020-bleurt} and COMET \citep{rei-etal-2020-comet} are supervised neural metrics fine-tuned on human evaluation. We employ the \textbf{BLEURT20} and \textbf{COMET-22} for reference-based metrics, and \textbf{COMET-QE} for the reference-less metric.
\textbf{UniTE} \citep{wan-etal-2022-unite} is also a learnt metric that evaluates MT outputs combining three different evaluation scenarios. We also adopt \textbf{UniTE-src} for comparing reference-less metrics. \textbf{MetricX-XXL} \citep{juraska-etal-2023-metricx} is a large-scale multi-task metric that fine-tunes LLM checkpoints using diverse human feedback data. For reference-less metrics, we also reproduce \textbf{MaTESe-QE} \citep{perrella-etal-2022-matese}, a metric leveraging transformer-based multilingual encoders to identify error spans in translations. 

\paragraph{Large Language Models} For prioprietary models, we use the OpenAI API to experiment with \textbf{GPT-3.5-Turbo}\footnote{We use the 0613 OpenAI model version. Due to budget constraints, we verify EAPrompt on GPT-4 with a limited number of samples. Please refer to Appendix~\ref{sec:appendix_gpt4} for more details.}. We also experiment with an inhouse human-aligned Llama2-70B series model \cite{touvron2023llama} fine-tuned with multilingual translation data, noted as "\textbf{Llama2-70b-Chat}" in experimental results. We also use a high-quality sparse mixture-of-experts model, Mixtral-8x7b \citep{jiang2024mixtral}. We use a state-of-the-art checkpoint \textbf{Mixtral-8x7b-Instruct} which has been optimised through supervised fine-tuning and direct preference optimisation to follow instructions.

\subsection{Prompts for LLM evaluators}

For GEMBA Prompting, we adopt the GEMBA-DA variant as suggested by \citep{kocmi-federmann-2023-large}, given its widespread usage and superior performance across three language pairs \citep{kocmi-federmann-2023-gemba}. 

For Error Analysis Prompting (EAPrompt), we conduct a comparison of various prompting strategies of EAPrompt in \S\ref{sec:ablation}, and use the best-performing variant for other experiments. We show the detailed prompt contexts in Appendix~\ref{sec:appendix_testprompt}. 

\subsection{Inference Costs for EAPrompt}

It is importatant to note that using a more complex prompting strategy may lead to higher inference costs. In EAPrompt, there are two steps that may require querying LLM for each translation:

\paragraph{Identifying Errors} This step generates a maximum of 256 tokens in total and may terminate early if all errors are listed.

\paragraph{Counting the Number of Errors} This step can be implemented in two ways. If querying LLM, it outputs only two numbers with a fixed output format, costing less than 10 tokens. Alternatively, if a regular expression matching algorithm is employed, this step does not require inference, thus reducing costs. As described in our experimental results later \ref{sec:repr_exp} these two approaches have minimal influence on the final performance.

\subsection{Experimental Results} \label{sec:mainres}

\begin{table*}[ht]
\centering
\small
\begin{tabular}{llccccc}
\toprule[0.5mm]
 & & & \textbf{System-Level Acc.} & \multicolumn{3}{c}{\textbf{Segment-Level Acc*}} \\
 \cmidrule(lr){4-4} \cmidrule(lr){5-7}
 \textbf{Models} & \textbf{Metrics} / \textbf{Prompts} & \textbf{Ref?} & \textbf{All (3 LPs)} & \textbf{En-De} & \textbf{En-Ru} & \textbf{Zh-En} \\\midrule
\multirow{7}{*}{\textbf{Baselines}}
 & MetricsX-XXL & \cmark & 85.0 & \underline{60.4} & \underline{60.6} & \underline{54.4} \\
 & BLEURT20 & \cmark & 84.7 & 56.8 & 54.0 & 48.9 \\
 & COMET22 & \cmark & 83.9 & 59.4 & 57.7 & 53.6 \\
 & UniTE & \cmark & 82.8 & 59.8 & 57.7 & 51.7 \\
\rowcolor{gray!16} \cellcolor{white} & COMET-QE & \xmark & 78.1 & 55.5 & 53.4 & 48.3\\
\rowcolor{gray!16} \cellcolor{white} & UniTE-src & \xmark & 75.9 & 58.2 & 55.4 & 50.8 \\
\rowcolor{gray!16} \cellcolor{white} & MaTESe-QE & \xmark & 74.8 & 57.2 & 49.9 & 49.4 \\\midrule
\multirow{4}{*}{\textbf{Llama2-70b-Chat}}
 & GEMBA & \cmark & 74.1 & 53.7 & 48.8 & 45.4 \\
 & EAPrompt & \cmark & 85.4 \textcolor{red}{(+11.3)} & 55.2\textcolor{red} {(+1.5)} & 51.4\dag \textcolor{red}{(+2.6)} & \textbf{50.2\dag \textcolor{red}{(+4.8)}} \\
 & \cellcolor{gray!16}GEMBA & \cellcolor{gray!16}\xmark & \cellcolor{gray!16}72.6 & \cellcolor{gray!16}54.1 & \cellcolor{gray!16}47.8 & \cellcolor{gray!16}45.0 \\
\rowcolor{gray!16} \cellcolor{white} & EAPrompt & \xmark & 85.8 \textcolor{red}{(+13.2)} & 55.0\dag \textcolor{red}{(+0.9)} & 51.6\dag \textcolor{red}{(+3.8)} & 49.3\dag \textcolor{red}{(+4.3)} \\\midrule
\multirow{4}{*}{\textbf{Mixtral-8x7b-Instruct}}
 & GEMBA & \cmark & 69.7 & 54.8 & 48.3 & 46.7 \\
 & EAPrompt & \cmark & 84.0 \textcolor{red}{(+14.3)} & 53.8 \textcolor{blue}{(-1.0)} & 50.6\dag \textcolor{red}{(+2.3)} & 48.2\dag \textcolor{red}{(+1.5)} \\
 & \cellcolor{gray!16}GEMBA & \cellcolor{gray!16}\xmark & \cellcolor{gray!16}74.1 & \cellcolor{gray!16}54.8 & \cellcolor{gray!16}47.5 & \cellcolor{gray!16}46.2 \\
\rowcolor{gray!16} \cellcolor{white} & EAPrompt & \xmark & 82.5 \textcolor{red}{(+8.4)} & 54.1 \textcolor{blue}{(-0.7)} & 49.9\dag \textcolor{red}{(+2.4)} & 48.3\dag \textcolor{red}{(+1.1)} \\\midrule
\multirow{4}{*}{\textbf{GPT-3.5-Turbo}}
 & GEMBA & \cmark & 86.5 & 55.2 & 49.5 & 48.2 \\
 & EAPrompt & \cmark & \textbf{\underline{91.2} \textcolor{red}{(+4.7)}} & \textbf{56.7\dag \textcolor{red}{(+1.5)}} & 53.3\dag \textcolor{red}{(+3.8)} & 50.0\dag \textcolor{red}{(+1.8)} \\
 & \cellcolor{gray!16}GEMBA & \cellcolor{gray!16}\xmark & \cellcolor{gray!16}86.9 & \cellcolor{gray!16}54.7 & \cellcolor{gray!16}50.0 & \cellcolor{gray!16}47.6 \\
 \rowcolor{gray!16} \cellcolor{white} & EAPrompt & \xmark & 89.4 \textcolor{red}{(+2.5)} & 55.7\dag \textcolor{red}{(+1.0)} & \textbf{53.4}\dag \textcolor{red}{(+3.4)} & 48.8\dag \textcolor{red}{(+1.2)} \\

\bottomrule[0.5mm]
\end{tabular}
\caption{\textbf{The performance of metrics using pairwise accuracy} (\%) \textbf{at the system level and pairwise accuracy with tie calibration} (\%) \textbf{at the segment level}. All results are compared with human-annotated MQM scores. The best results among the same model are highlighted in \textbf{bold}. The best results among all metrics are \underline{underlined}. "\dag" denotes cases where one metric is significantly better than the other.}
\label{tab:mainres_new}
\end{table*}

We compute system\&segment level performance of EAPrompt with LLMs in Table~\ref{tab:mainres_new}. We see that: 

\paragraph{(i) At the system level, EAPrompt empowers GPT-3.5-Turbo to surpass all other metrics and achieves state-of-the-art performance.} 

Consistent with the findings of \citet{kocmi-federmann-2023-large}, LLMs achieve state-of-the-art performance across all three language pairs at the system level, significantly outperforming traditional metrics ("\textbf{Baselines}") by a large margin. 

Remarkably, when prompting all LLMs with EAPrompt, the performance notably surpasses GEMBA at the system level, achieving the highest pairwise accuracy of 91.2\% on \textbf{GPT-3.5-Turbo}, thus establishing a new SOTA.

\paragraph{(ii) At the segment level, EAPrompt outperforms GEMBA in 8 out of 9 tested scenarios.}

At the segment level, despite previous findings by \citet{kocmi-federmann-2023-large} regarding the weak correlation between LLMs as evaluators and human judgments, prompting with EAPrompt addresses this drawback of LLM evaluators, outperforming GEMBA's performance on nearly all tested LLMs and language pairs by a significant margin. The best segment-level results are achieved by \textbf{GPT-3.5-Turbo} for En-De (56.7) and En-Ru (53.4), and by \textbf{Llama2-70b-Chat} for Zh-En (50.2). This validates the effectiveness of our EAPrompt.

The only exception of the result is observed for En-De \textbf{Mixtral-8x7b-Instruct}, where the segment-level accuracy is lower than GEMBA by 1.0. This discrepancy might be attributed to the limited capability of identifying translation errors in En-De language pair. Another notable finding is that prompting with LLMs, both with GEMBA and EAPrompt, fails to surpass current best metrics ("\textbf{Baselines}") at the segment level. This could be because these baseline metrics have been fine-tuned using extensive translation and human evaluation datasets, while the LLMs employed in our experiment are versatile models guided by few-shot prompts.

\paragraph{(iii) EAPrompt enhances the performance of LLMs as translation evaluators in reference-less scenarios.}

Our main findings remain consistent with both reference-based and reference-less settings (indicated by "\cmark" and "\xmark" in \textbf{Ref?}, respectively), where EAPrompt continues to outperform GEMBA across all three tested LLMs at the system level, and in 8 out of 9 scenarios at the segment level. The improvement is slightly lower compared to scenarios with referenced signals.

These results underscore the impressive cross-lingual capabilities of LLMs and their suitability for quality estimation under EAPrompt, even in the absence of reference translations, which poses a significant challenge on MT evaluation.

\begin{table*}[ht]
\centering
\Large
\resizebox{\linewidth}{!}{
\begin{tabular}{ccccccccccccc}
\toprule[0.5mm] 
\textbf{Prompt} & \multicolumn{2}{c}{\textbf{Demo of Errors}} & \multicolumn{2}{c}{\textbf{Type of Queries}} & \multicolumn{4}{c}{\textbf{Mixtral-8x7b-Instruct}} & \multicolumn{4}{c}{\textbf{Llama2-70b-Chat}} \\
\cmidrule(lr){2-3} \cmidrule(lr){4-5} \cmidrule(lr){6-9} \cmidrule(lr){10-13}
  & Detailed & Itemized & 1-step & 2-step & All (3 LPs) & En-De & En-Ru & Zh-En & All (3 LPs) & En-De & En-Ru & Zh-En \\ \midrule
\textbf{GEMBA} & - & - & - & - & 69.7 & \textbf{54.8} & 48.3 & 46.7 & 74.1 & 53.7 & 48.8 & 45.4 \\ \cmidrule(lr){1-13}
\multirow{4}{*}{\textbf{EAPrompt}} & \cmark & & \cmark & & 75.2 & 53.4 & 50.0 & 45.0 & 62.0 & 53.7 & 47.0 & 47.8 \\
 & \cmark & & & \cmark & 75.5 & 53.4 & 47.9 & 45.5 & 84.7 & 53.5 & 46.9 & 47.5 \\
 & & \cmark & \cmark & & 60.2 & 53.4 & 45.1 & 45.6 & 56.9 & 53.7 & 48.4 & \textbf{50.2} \\
 & & \cmark & & \cmark & \textbf{84.0} & 53.7 & \textbf{50.6} & \textbf{48.2} & \textbf{85.4} & \textbf{55.2} & \textbf{51.4} & \textbf{50.2} \\

\bottomrule[0.5mm]
\end{tabular}}
\caption{\textbf{Comparison of the system level} ("All (3 LPs)") \textbf{and segment level} ("En-De", "En-Ru", "Zh-En") \textbf{performance of LLMs with different variants of prompts for EAPrompt}. We compare itemized or detailed responses to demonstrate identified errors. We also compare the instructions, whether separated into two queries (marked as "2-step", one for identifying errors and another for scoring) or combined into a single query (marked as "1-step"). The
best results among all prompt variants are highlighted in \textbf{bold}.} 
 \label{tab:prompts_selection_new}
\end{table*}

\subsection{Ablation Study of Prompt Variants}  \label{sec:ablation}

Given the crucial significance of the prompt design, we investigate several versions of in-context prompt contexts and present an analysis in Table~\ref{tab:prompts_selection_new}. The prompt contexts used in our experiment are detailed in Appendix~\ref{sec:appendix_testprompt}. Due to budget constraints, we utilize two LLMs, \textbf{Mixtral-8x7b-Instruct} and \textbf{Llama2-70b-Chat}, as the test bed for this ablation study. Our findings indicate that:

\paragraph{(i) Itemized error demonstration is superior to detailed illustration.} 
We assume that when identifying translation errors, providing detailed descriptions may impede the LLM's capability to accurately identify errors and count the number of them. As illustrated in the "\textbf{Demo of Errors}" column, employing itemized error demonstrations instead of detailed paragraphs yields improved performance at both the system and segment levels for both tested LLMs.

In our initial study, we observed that generating excessively detailed responses could lead to incorrect error counting or misclassification of error severity. Therefore, it is recommended to employ clear and concise error descriptions in a format that is easily processed and comprehended by LLMs.

\paragraph{(ii) Separating the scoring process from error identification with two queries will enhance the performance of LLMs as translation evaluators.} Another consideration in prompt design is the division of the evaluation process into error identification and error counting. As depicted in the "\textbf{Type of Queries}" column, it is evident that the performance of using a single prompting step is considerably lower than that of employing a  2-step prompting approach. This may be because separating the scoring process allows LLMs to concentrate on a single task in each query, thereby facilitating more accurate judgments and reducing the likelihood of incorrectly counting the number of errors.

\paragraph{(iii) Among the prompting strategies, EAPrompt appears to be more suitable for the LLMs as translation evaluators.} When compared with GEMBA prompting strategies, the EAPrompt variant featuring a 2-step separated prompting approach and itemized response achieves superior performance in enhancing LLMs' effectiveness as translation evaluators. Consequently, we recommend employing this particular variant for LLMs as translation evaluators.

\section{Analysis}

\subsection{EAPrompt aligns with human judgment through similar distribution of major and minor errors across most LLMs} \label{sec:dist_shift}

\begin{figure*}[t]

\includegraphics[scale=0.36]{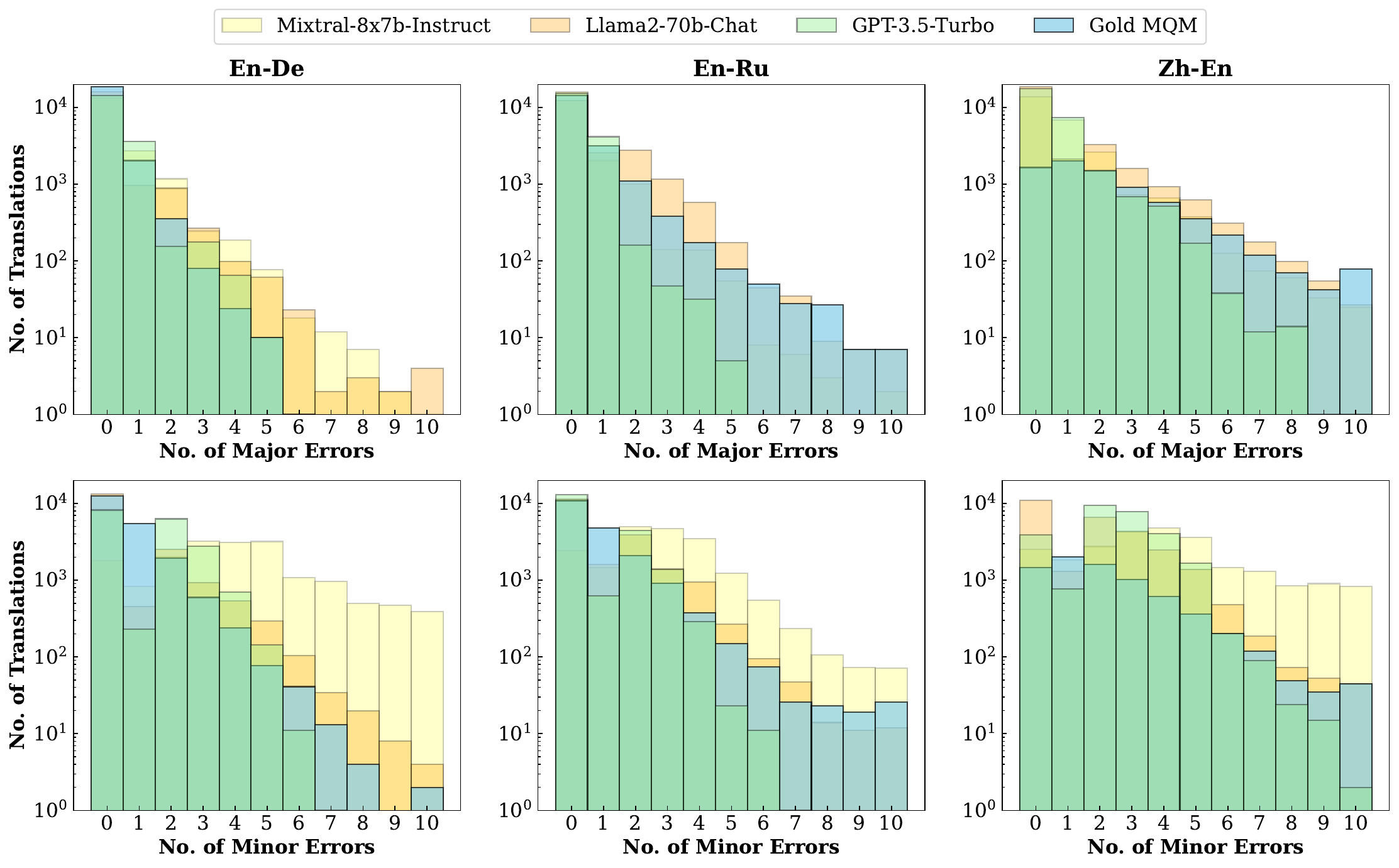}
\centering
\caption{\textbf{Distribution of identified error counts} across various LLMs and human evaluation (MQM), for the language pairs En-De, En-Ru and Zh-En, repectively.}
\label{fig:dist_shift}
\end{figure*}

To investigate can LLMs align with gold human judgement MQM through similar distributions of major and minor errors, we present the error distribution across various test scenarios in Figure~\ref{fig:dist_shift}. 

We can see that, for major errors, all tested LLMs exhibit distributions that closely resemble MQM. Regarding minor errors, Mixtral-8x7b-Instruct appears to produce a slightly higher frequency of such errors compared to other LLMs, while the distribution of other LLMs remains consistent with MQM. This observation further validates the efficacy of EAPrompt. 

This finding provides valuable insights into enhancing the reliability of LLMs as translation evaluators. It suggests a potential focus on guiding LLMs to more accurately identify minor errors, such as clarifying the specific categories and severity of minor errors.

\subsection{EAPrompt empowers LLMs to distinguish major errors from minor ones} \label{sec:major_weight_adjust}

\begin{figure}[t]

\includegraphics[scale=0.5]{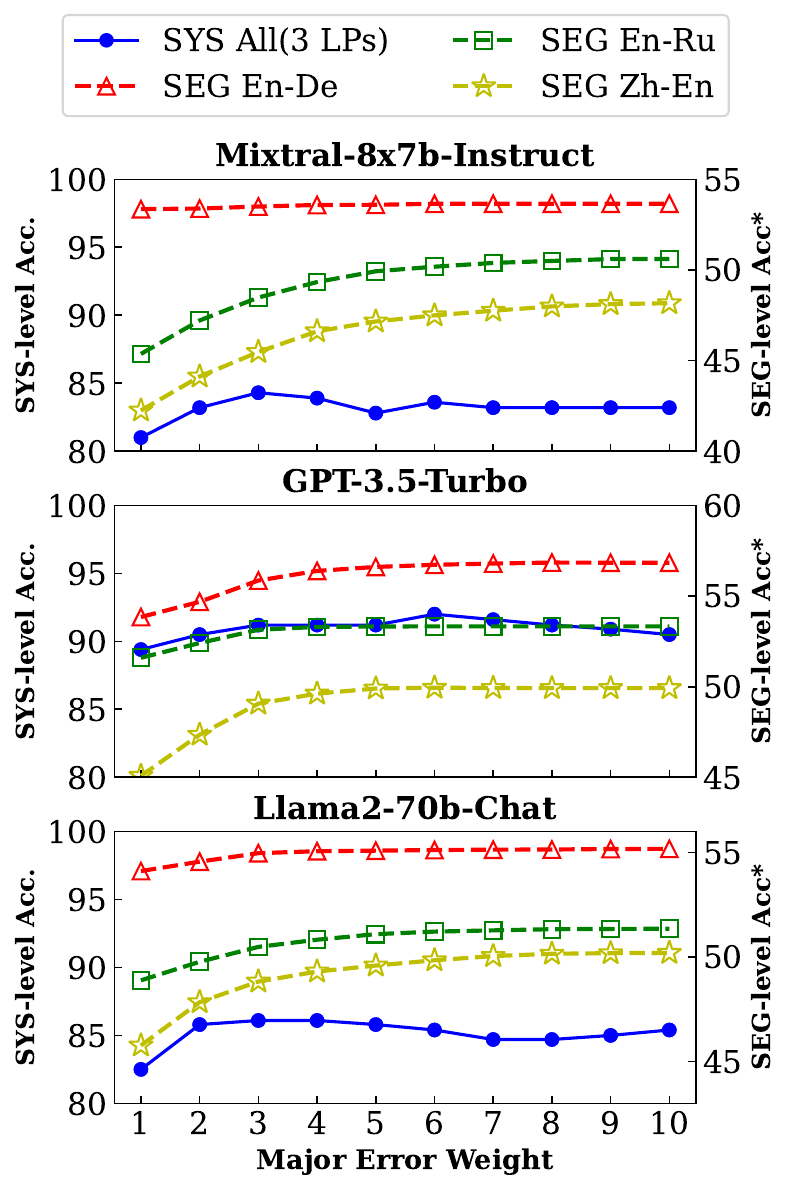}
\centering
\caption{\textbf{Effect of varying major error weight} ($w_{\text{major}}$) on EAPrompt across different LLMs at both system and segment levels.}
\label{fig:weight_influence}
\end{figure}

A potential concern on EAPrompt is whether this technique can prompt LLMs to distinguish major errors from minor ones. To address this concern, we adjust 
the weight assigned to major errors ($w_{\text{major}}$) in the score computation process outlined in \S\ref{sec:postprocess}. We visualize the impact of this adjustment on both the system and segment-level performance in Figure~\ref{fig:weight_influence}. If the metric effectively distinguishes major errors from minor ones, we anticipate a noticeable performance decrease when the weight of major errors $w_{\text{major}}$ approaches that of minor errors ($w_{\text{minor}}$ = 1 in this study). 

Our findings reveal that for all three LLMs tested, adjusting $w_{\text{major}} < 3$ results in a substantial performance decline, indicating that prompting error analysis with all tested LLMs possesses the ability to discriminate major errors from minor ones.

Another noteworthy observation from this analysis is that when $w_{\text{major}} \geq 5$, both the system-level and segment level-accuracies exhibit minimal fluctuation, suggesting that the performance of EAPrompt remains nearly unaffected by this latent variable during score computation. 

\subsection{EAPrompt optimizes inference costs by utilizing regular expressions instead of counting queries} \label{sec:repr_exp}

\begin{table*}[ht]
\centering
\small
\begin{tabular}{lccccc}
\toprule[0.5mm]
 & & \textbf{System-Level Acc.} & \multicolumn{3}{c}{\textbf{Segment-Level Acc*}} \\
 \cmidrule(lr){3-3} \cmidrule(lr){4-6}
 \textbf{Models} & \textbf{Repr?} & \textbf{All (3 LPs)} & \textbf{En-De} & \textbf{En-Ru} & \textbf{Zh-En}\\\midrule
\multirow{2}{*}{\textbf{Llama2-70b-Chat}}
 & \cmark & 85.0 & 55.6 & 51.5 & 50.4 \\
 & \cellcolor{gray!16}\xmark & \cellcolor{gray!16}85.4 & \cellcolor{gray!16}55.2 & \cellcolor{gray!16}51.4 & \cellcolor{gray!16}50.2 \\\midrule
\multirow{2}{*}{\textbf{Mixtral-8x7b-Instruct}}
 & \cmark & 82.8 & 53.7 & 50.9 & 47.6 \\
 & \cellcolor{gray!16}\xmark & \cellcolor{gray!16}84.0 & \cellcolor{gray!16}53.7 & \cellcolor{gray!16}50.6 & \cellcolor{gray!16}48.2 \\\midrule
\multirow{2}{*}{\textbf{GPT-3.5-Turbo}}
 & \cmark & 90.1 & 56.8 & 53.9 & 50.0 \\
 & \cellcolor{gray!16}\xmark & \cellcolor{gray!16}91.2 & \cellcolor{gray!16}56.7 & \cellcolor{gray!16}53.3 & \cellcolor{gray!16}50.0  \\
\bottomrule[0.5mm]
\end{tabular}
\caption{\textbf{Performance comparison of EAPrompt between utilizing the Regular Expression Matching strategy} ("\cmark" in \textbf{Repr?}) and the counting query strategy ("\xmark" in \textbf{Repr?}) across various LLMs.}
\label{tab:regular_expr}
\end{table*}

Since EAPrompt adopts a two-step prompting strategy, one related question is: can we simplify the query process to reduce inference costs? One potential approach involves substituting the scoring query step with an algorithm that identifies major and minor errors using regular expressions (\textbf{Repr}) to detect bullet points or initial numbers. A detailed description of the \textbf{Repr} matching strategy is provided in the Appendix. The analysis, as depicted in Table~\ref{tab:regular_expr}, indicates that employing Repr matching strategy, as opposed to the original query for counting errors (indicated by "\xmark" in \textbf{Repr?}), yields minimal performance variation at both system and segment levels. Thus, if inference costs are a concern for this metric, substituting the second query step of EAPrompt with regular expressions could be a viable option. Note that for different LLMs, a tailored regular expression pattern may be necessary to encompass various response structures.

\subsection{Case Study}

We discuss potential issues encountered by LLMs and their corresponding solutions in Appendix~\ref{sec:casestudy}, including invalid responses, input order bias, etc. We aim to provide insights that should be considered when utilizing LLMs as translation evaluators.

\section{Related Work}

\paragraph{Translation Evaluation Metrics} MT Evaluation metrics are of crucial importance to the development of MT systems \citep{freitag-etal-2022-results}. Studies have shown that traditional surface-based metrics such as BLEU \citep{papineni-etal-2002-bleu} are no longer suitable for evaluating high-quality MT systems \citep{mathur-etal-2020-tangled}. Modern metrics like COMET \citep{rei-etal-2020-comet}, MetricsX-XXL \citep{juraska-etal-2023-metricx}, BLEURT \citep{sellam-etal-2020-bleurt}, and UniTE \citep{wan-etal-2022-unite} leverage human evaluations and high-quality translations for training. While these metrics achieve strong correlation with human judgements such as MQM \citep{freitag-etal-2021-experts}, there is a growing demand for explainability in their evaluation. Despite progress, recent research struggles to strike a balance between the reliability and explainability of these metrics \citep{lu-etal-2023-toward, xu-etal-2022-errors, perrella-etal-2022-matese}. In this work, we delve into the potential of LLMs for "human-like" translation evaluation, as they possess the capability to explicitly identify translation errors without further fine-tuning, which resembles the evaluation process of human.

\paragraph{LLMs as Evaluators} LLMs refers to language models with hundreds of billion of parameters which are trained on massive textual data \citep{evaluation-survey-2024, zhao2023survey}. Since the emergence of ChatGPT, LLMs have shown remarkable proficiency across various NLP tasks \citep{achiam2023gpt, touvron2023llama}. A prevalent application of LLMs is harnessing them as evaluators for assessing the performance of Chatbots \citep{zheng2023judging} or some self-criticize/ improvement procedures~\cite{zhang2024intention, valmeekam2023investigating}. Recent studies also show LLM's efficacy in evaluating NLG tasks like summarization and dialogue generation through multi-step prompting \citep{liu-etal-2023-g}. GEMBA \citep{kocmi-federmann-2023-large} is the pioneering effort in utilizing LLMs as translation evaluators via a zero-shot prompting approach with GPT models. In this work, EAPrompt innovatively combines the basic ideas of error analysis~\citep{lu-etal-2023-toward} and chain-of-thought~\citep{wei2022chain} to prompt LLMs for achieving human-like translation evaluation.

Subsequent work follows our work to further explore the potential of LLMs as translation evaluators. AutoMQM \citep{fernandes-etal-2023-devil} parallels our approach, utilizing PaLM-2 model \citep{anil2023palm} as the testbed. GEMBA-MQM \citep{kocmi-federmann-2023-gemba} further improves EAPrompt by employing a  few-shot prompting technique using GPT-4, making this approach universally applicable across languages. Another line of research focuses on fine-tuning LLMs to accurately predict error spans in translations. For instance, InstructScore \citep{xu-etal-2023-instructscore} fine-tunes a Llama model \citep{touvron2023llama1}, while XCOMET \citep{guerreiro2023xcomet} scales from COMETKiwi \citep{rei-etal-2023-scaling} to achieve this goal.

\section{Conclusion}
\label{sec:conclusion}
In this paper, we explore the potential of LLMs as a metric for evaluating translations. We design a novel one-shot prompting strategy EAPrompt based on chain-of-thought and error analysis, and show that this strategy significantly improves the evaluation performance on both the system and segment levels. We compare different EAPrompt variants and ultimately opt for a 2-step prompting approach with itemized error demonstrations. Further analysis confirms EAPrompt's proficiency in error identification and its alignment with the commonly accepted human evaluations MQM. 
In future work, we would like to experiment with a broader range of LLMs~\cite{wmt19,iwslt21,wmt22,vegamt}, to make our conclusion more convincing. 
Lastly, it will be interesting to test the capabilities of LLMs for other MT-related tasks, such as grammatical error correction and automatic post-editing \cite{wu2023chatgpt, vidal-etal-2022-automatic}.

\section*{Limitations}

The limitations of this work are three-fold:

\begin{itemize}
    \item Potential Test Data Contamination: Although we utilized WMT22 to minimize the risk of test set leakage in the training data of LLMs, it is still possible that some contamination from the test data remains. Therefore, future researchers utilizing these datasets should be cautious and carefully address this issue, as it may affect the availability of the test set for comparison purposes.
    \item Budget Constraints: Due to limited resources, we were unable to explore more prompt choices comprehensively in our research. The findings presented in this study only reflect our initial experiments. We leave the impact of different prompt choices for further investigation.
    \item Limited Range of LLMs Tested: In this study, we focused on evaluating a limited number of LLMs that we believed possessed potential and capability as translation evaluators. However, it is important to note that not all existing LLMs can necessarily serve as reliable evaluators under the EAPrompt approach. Future research could explore and experiment with a broader range of LLMs, examining their effectiveness and assessing their suitability as evaluators.
\end{itemize}

\section*{Ethics Statement}

We take ethical considerations very seriously, and strictly adhere to the Code of Ethics. All procedures performed in this study are in accordance with the ethical standards. This paper focuses on evaluating the capabilities of LLM as a translation evaluator. Our proposed approach, EAPrompt, does not include statements that induce the model to generate harmful information. Additionally, this method solely extracts and processes the numerical scores from the model's response, thereby further mitigating the potential risks. Both the datasets and models used in this paper are publicly available and have been widely adopted by researchers. Our model will not learn from user inputs or cause potential risks to the NLP community. We ensure that the findings and conclusions of this paper are reported accurately and objectively. Informed consent was obtained from all individual participants included in this study.

\section*{Acknowledgments}
We thank the anonymous reviewers and the area chair for their insightful comments and suggestions. This work was supported in part by the National Natural Science Foundation of China under Grant 61973083, and in part by the Shenzhen Science and Technology Program JCYJ20210324121213036. 

\bibliography{anthology,arxiv_version}

\appendix

\section{Description of MQM} \label{appendix:MQM}
Multidimensional Quality Metric (MQM) is a human evaluation framework commonly used in WMT metrics shared tasks as the golden standard \citep{freitag-etal-2021-experts, freitag-etal-2023-results}. It is developed to evaluate and categorize errors in translations. The annotations from human experts are open-sourced and available from WMT22 metrics shared tasks for scientific research \citep{freitag-etal-2022-results}. 

In WMT22, MQM annotations for En-De and Zh-En were sponsored and executed by Google, using 11 professional translators (7 for En-De, 4 for Zh-En). The annotations for En-Ru were provided by Unbabel who utilized 4 professional, native language annotators with ample translation experience. They have access to the full document context. 

About the inter-rater agreement. In MQM, each segment is annotated by 2 or 3 annotators. The final segment-level score is an average over scores from all annotators. As depicted in \citep{freitag-etal-2021-experts}, the pairwise inter-rater agreement is about 0.584 for En-De, and 0.412 for Zh-En, which is significantly better than other evaluation protocols such as Scalar Quality Metric and Direct Assessment.

In this paper, EAPrompt emulates MQM to identify major and minor errors, providing insightful explanations for the translation. Table~\ref{tab:case_info} shows an example annotated through MQM framework.

\begin{table*}[ht]
\centering
\small
\begin{tabular}{ll}
\toprule[0.5mm] 
\textbf{System} &  Online-A.en\\
\textbf{Domain} &  conversational\\
\textbf{Doc\_id}  & 1 \\
\textbf{Seg\_id}  & 6 \\\midrule
\textbf{Source(Zh)} & \begin{CJK}{UTF8}{gbsn}请问，订单情况现在是什么样？\end{CJK} \\
\textbf{Reference(En)} & May I ask what the status of the order is now? \\
\textbf{Translation(En)} & Please ask, what is the order situation now? \\\midrule
\textbf{Major Error(s)} & "Please ask" - Accuracy/Mistranslation\\
\textbf{Minor Error(s)} & "situation" - Style/Awkward \\
\bottomrule[0.5mm]
\end{tabular}
\caption{\textbf{An example of MQM}, comprising information of the test sample along with human-annotated errors.}
\label{tab:case_info}
\end{table*}

\section{Post-processing of EAPrompt} \label{sec:appendix_post}

As described in \S\ref{sec:postprocess}, we treat $w_{\text{major}}$ as a latent variable within EAPrompt.
In our experiments, we select this latent variable with the best averaging performance for each LLMs denoted as $w_{\text{major}}^*$. The value was reported in Table~\ref{tab:error_weight}.

\begin{table}[ht]
\centering
\small
\begin{tabular}{lc}
\toprule[0.5mm] 
\textbf{Model} & \textbf{$w_{\text{major}}^*$} \\ \midrule
GPT-3.5-Turbo & 6 \\
Llama2-70b-Chat & 10 \\
Mixtral-8x7b-Instruct & 10 \\
\bottomrule[0.5mm]
\end{tabular}
\caption{\textbf{Optimal values of $w_{\text{major}}^*$ for each LLM}. To ensure fair comparison, we maintain this variable constant across all tested scenarios for every LLM.}
 \label{tab:error_weight}
\end{table}

\section{Prompt Contexts of EAPrompt} \label{sec:appendix_testprompt}
\begin{figure*}[ht]
\includegraphics[scale=0.67]{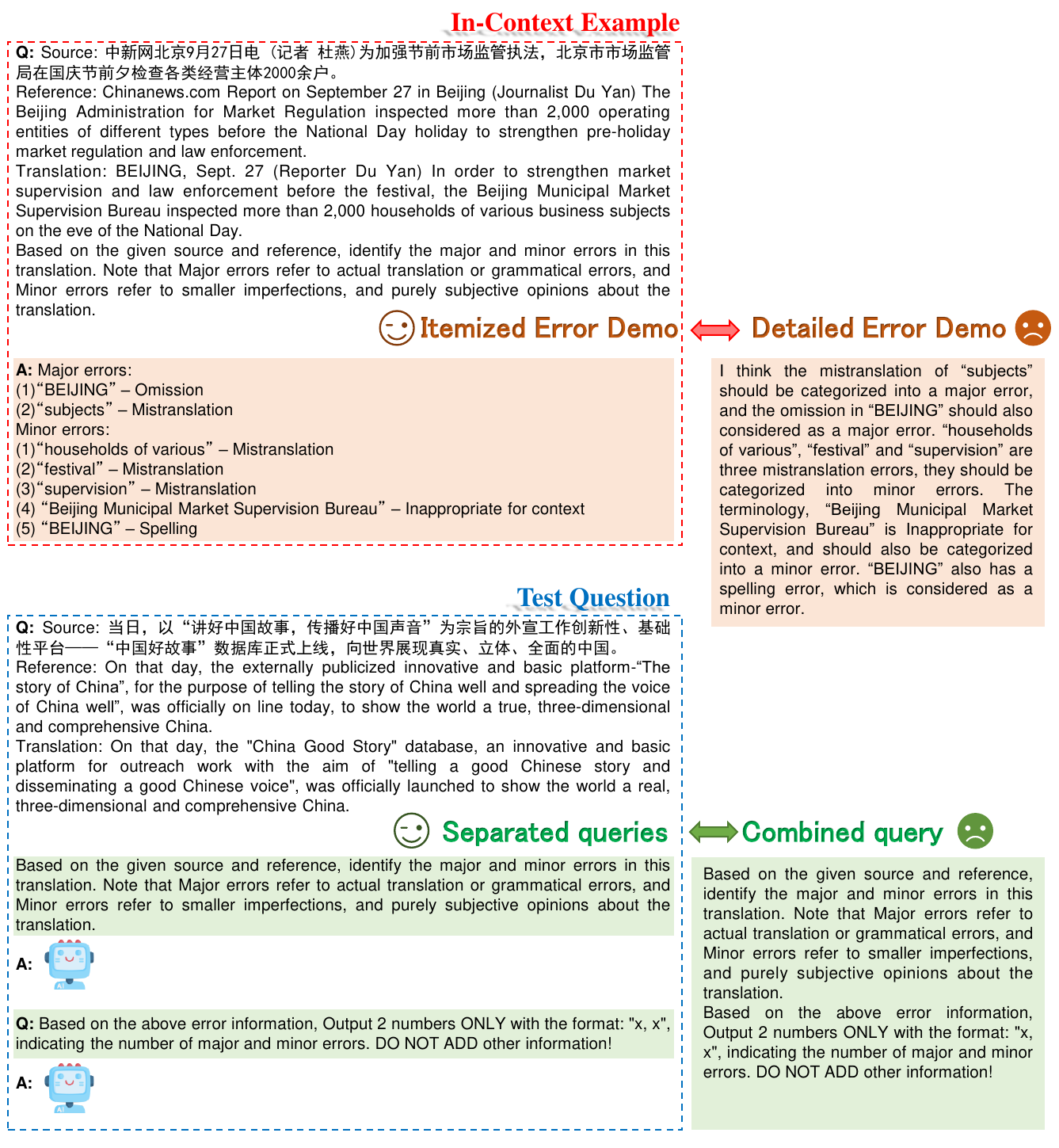}
\centering
\caption{\textbf{The prompt contexts employed in EAPrompt}. We present itemized/detailed responses for error demonstrations and separated/combined instructions for different types of queries.}
\label{fig:bestprompt}
\end{figure*}

Figure~\ref{fig:bestprompt} provides the prompt contexts implemented in EAPrompt, along with the detailed error demonstration and combined query instruction discussed in \S\ref{sec:ablation} for reproduction of our experiments.

\section{Counting Errors using Regular Expressions Matching}

\begin{figure*}[ht]
\includegraphics[scale=0.47]{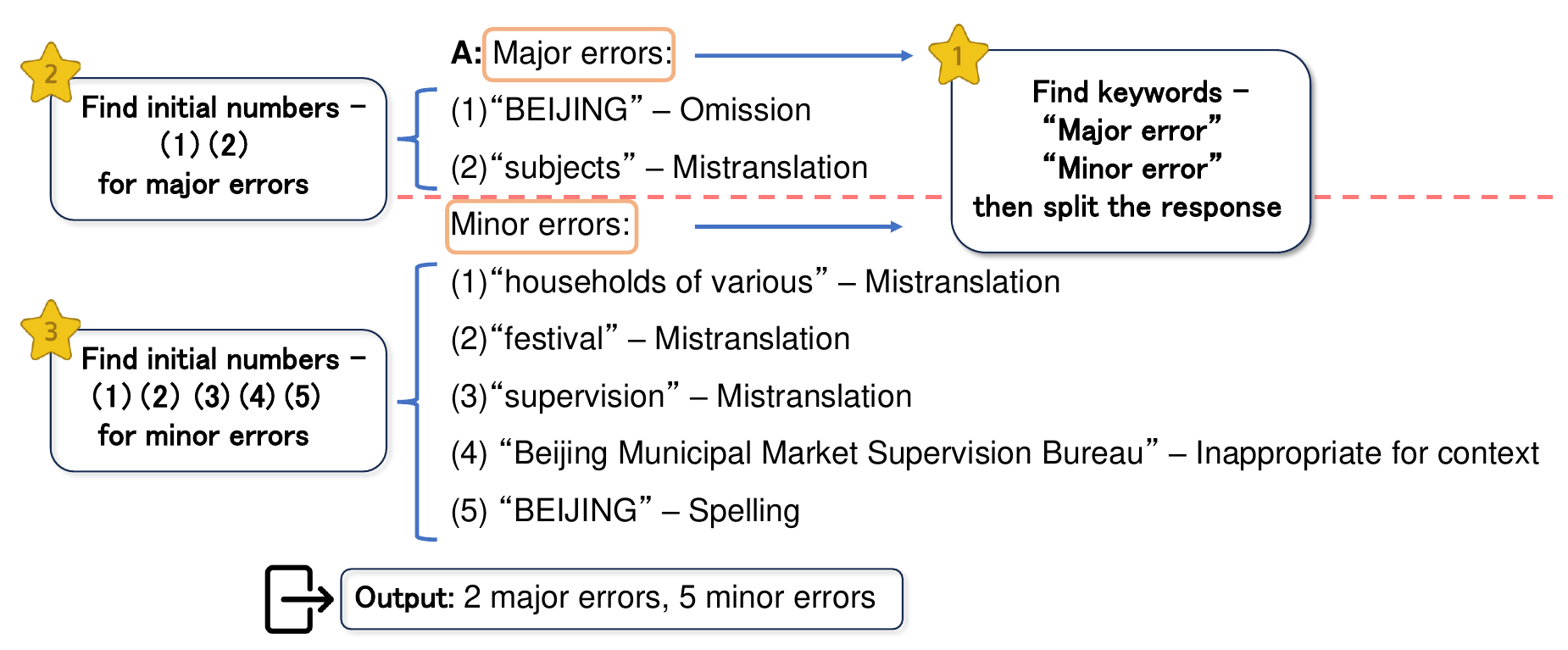}
\centering
\caption{\textbf{The regular expression matching strategy} utilized in \S\ref{sec:repr_exp} to automatically count the number of major and minor errors in the LLM response.}
\label{fig:expr}
\end{figure*}

In Figure~\ref{fig:expr}, we present an overview of our error-matching strategy utilized in \S\ref{sec:repr_exp} to automatically identify the number of major and minor errors. The procedure can be listed as follows:

\begin{itemize}
    \item[1.] Locate "major error" and "minor error" within the response, then segment the response accordingly.
    \item[2.] Utilize Regular Expression matching to identify the initial numbers of major and minor errors. For implementation, we include three different initial number formats: "1.", "1)" and "(1)" (using "1" as an example);
    \item[3.] Record the number of major and minor errors.
\end{itemize}

\section{Additional Results} 

\subsection{Results on GPT-4} \label{sec:appendix_gpt4}

In order to further verify the effect of EAPrompt on state-of-the-art LLMs, we test our approach on GPT-4 as an initial experiment, with a subset of 600 samples, due to budget constraints. Results are shown in Table~\ref{tab:gpt-4res}. We can see that on powerful GPT-4, EAPrompt shows a slight performance advantage over GEMBA-DA, outperforming it by 1.05 points.

\begin{table}[ht]
\centering
\small
\begin{tabular}{lccc}
\toprule[0.5mm] 
\textbf{Model} & \textbf{GEMBA} & \textbf{EAPrompt} & \textbf{$\delta$} \\\midrule
GPT-4 & 44.92 & 45.97 & +1.05 \\\bottomrule[0.5mm]
\end{tabular}
\caption{Segment Level Kendall Correlation on WMT22 Zh-En using GPT-4 with a subset of 600 samples.}
 \label{tab:gpt-4res}
\end{table}

\subsection{Results at the system-level}

\begin{table*}[ht]
\centering
\small
\begin{tabular}{llccccccc}
\toprule[0.5mm]
 & & & \multicolumn{2}{c}{\textbf{En-De}} & \multicolumn{2}{c}{\textbf{En-Ru}} & \multicolumn{2}{c}{\textbf{Zh-En}} \\
 \cmidrule(lr){4-5} \cmidrule(lr){6-7} \cmidrule{8-9}
 \textbf{Models} & \textbf{Prompts} & \textbf{Ref?} & \textbf{$\rho$} & \textbf{Acc.} & \textbf{$\rho$} & \textbf{Acc.} & \textbf{$\rho$} & \textbf{Acc.} \\\midrule
\multirow{4}{*}{\textbf{Llama2-70b-Chat}}
 & GEMBA & \cmark & 53.0 & 69.2 & 79.4 &	84.8 & 77.3 & 77.3 \\
 & EAPrompt & \cmark & \textbf{78.3} & 76.9 & 84.2 & 86.7 & \textbf{98.2} & \underline{\textbf{98.2}} \\
 & GEMBA & \xmark & 61.5 & 53.8 & 83.2 &	84.8 & 93.4 & 74.7 \\
 & EAPrompt & \xmark & 76.2 & \textbf{83.3} &	\textbf{85.7} & \textbf{88.6} & 97.6 & 84.6 \\\midrule
\multirow{4}{*}{\textbf{Mixtral-8x7b-Instruct}}
 & GEMBA & \cmark & 35.1 & 59.0 & 64.3 & 73.3 & 89.9 & 74.7 \\
 & EAPrompt & \cmark & 54.6 & 74.4 & \textbf{84.5} & \textbf{88.6} & 95.8 & \textbf{95.8} \\
 & GEMBA & \xmark & 70.7 & \textbf{82.1} & 83.0 & 85.7 & \textbf{98.0}	& 52.7 \\
 & EAPrompt & \xmark & \textbf{71.4} & 78.2 &	83.6 & 86.7 & 97.5 & 82.4 \\\midrule
\multirow{4}{*}{\textbf{GPT-3.5-Turbo}}
 & GEMBA & \cmark & 87.2 & 85.9 & 81.7 & 90.5 & \underline{\textbf{98.4}} & 82.4 \\
 & EAPrompt & \cmark & \underline{\textbf{92.1}} & \underline{\textbf{91.0}} & \underline{\textbf{92.0}} & \underline{\textbf{92.4}} & 98.3 & \textbf{92.3} \\
 & GEMBA & \xmark & 86.6 & 83.3 & 77.3 &	90.5 & 98.3 & 85.7 \\
 & EAPrompt & \xmark & 91.9 & 88.5 & 82.4 & 90.5 & 98.2 & 89.0 \\
\bottomrule[0.5mm]
\end{tabular}
\caption{\textbf{The performance of metrics for each language pair using pearson correlation ($\rho$) and pairwise accuracy (Acc.) at the system level}. All results are compared with human-annotated MQM scores. The best results among the same model are highlighted in \textbf{bold}. The best results among all metrics are \underline{underlined}.}.
\label{tab:additional_sys}
\end{table*}

To complement our main findings, we also compare our method, EAPrompt, with GEMBA using pearson correlation and pairwise accuracy. In addition to Table~\ref{tab:mainres_new}, which groups three language pairs together to present comprehensive and unbiased results, Table~\ref{tab:additional_sys} provides a system-level performance comparison across different language pairs for a more detailed analysis.

We can see that at the system level, across all LLMs and for all three language pairs, EAPrompt consistently outperforms GEMBA in most scenarios, aligning with the primary conclusions drawn in our paper. Additionally, we note a few cases where the two metrics exhibit slight discrepancies. This discrepancy, typically less than 1.0 point, may be attributed to Pearson correlation's sensitivity to low-quality MT systems \citep{mathur-etal-2020-tangled}, potentially leading to an unfair judgment compared to Pairwise Accuracy.

\subsection{Results at the segment-level}

\begin{table*}[ht]
\centering
\small
\begin{tabular}{llccccccc}
\toprule[0.5mm]
 & & & \multicolumn{2}{c}{\textbf{En-De}} & \multicolumn{2}{c}{\textbf{En-Ru}} & \multicolumn{2}{c}{\textbf{Zh-En}} \\
 \cmidrule(lr){4-5} \cmidrule(lr){6-7} \cmidrule{8-9}
 \textbf{Models} & \textbf{Prompts} & \textbf{Ref?} & \textbf{$\rho$} & \textbf{Acc*} & \textbf{$\rho$} & \textbf{Acc*} & \textbf{$\rho$} & \textbf{Acc*} \\\midrule
\multirow{4}{*}{\textbf{Llama2-70b-Chat}}
 & GEMBA & \cmark & 10.8 & 53.7 & 6.3 & 48.8 & 2.0 &	45.4 \\
 & EAPrompt & \cmark & \textbf{27.8}\dag & \textbf{55.2} & 28.1\dag & 51.4\dag & \textbf{39.1}\dag & \underline{\textbf{50.2}}\dag \\
 & GEMBA & \xmark & 4.3 & 54.1 & 5.0 & 47.8 & 1.2 & 45.0 \\
 & EAPrompt & \xmark & 26.2\dag & 55.0\dag &	\textbf{29.8}\dag & \textbf{51.6}\dag & 38.6\dag & 49.3\dag \\\midrule
\multirow{4}{*}{\textbf{Mixtral-8x7b-Instruct}}
 & GEMBA & \cmark & 23.8 & \textbf{54.8} & 6.4 & 48.3 & 2.5 &	46.7 \\
 & EAPrompt & \cmark & \textbf{30.1}\dag & 53.8 & \textbf{28.5}\dag & \textbf{50.6}\dag & \textbf{40.2}\dag & 48.2\dag \\
 & GEMBA & \xmark & 12.2 & \textbf{54.8} & 7.7 & 47.5 & 0.9 & 46.2 \\
 & EAPrompt & \xmark & 26.7\dag & 54.1 & 24.5\dag & 49.9\dag & 37.9\dag & \textbf{48.3}\dag \\\midrule
\multirow{4}{*}{\textbf{GPT-3.5-Turbo}}
 & GEMBA & \cmark & 28.6 & 55.2 & 21.0 & 49.5 & 24.8 & 48.2 \\
 & EAPrompt & \cmark & \underline{\textbf{39.9}}\dag & \underline{\textbf{56.7}}\dag & \underline{\textbf{34.6}}\dag & 53.3\dag & \underline{\textbf{41.9}}\dag & \textbf{50.0}\dag \\
 & GEMBA & \xmark & 29.6 & 54.7 & 24.3 &	50.0 & 24.9 & 47.6 \\
 & EAPrompt & \xmark & 33.3\dag & 55.7\dag &	30.2\dag & \underline{\textbf{53.4}}\dag & 36.2\dag & 48.8\dag \\
\bottomrule[0.5mm]
\end{tabular}
\caption{\textbf{The performance of metrics for each language pair using pearson correlation ($\rho$) and pairwise accuracy with ties (Acc*) at the segment level}. All results are compared with human-annotated MQM scores. The best results among the same model are highlighted in \textbf{bold}. The best results among all metrics are \underline{underlined}. "\dag" denotes cases where one metric is significantly better than the other.}.
\label{tab:additional_seg}
\end{table*}

We present more detailed results, including Pearson correlation, as shown in Table~\ref{tab:additional_seg}. We can observe that when using Pearson correlation as the meta-evaluation method, EAPrompt surpasses GEMBA by a significant margin across all scenarios considered in our experiments. This further validates the effectiveness and strong performance of EAPrompt compared to GEMBA.

\section{Case Study} \label{sec:casestudy}

\begin{figure*}[t]

\includegraphics[scale=0.54]{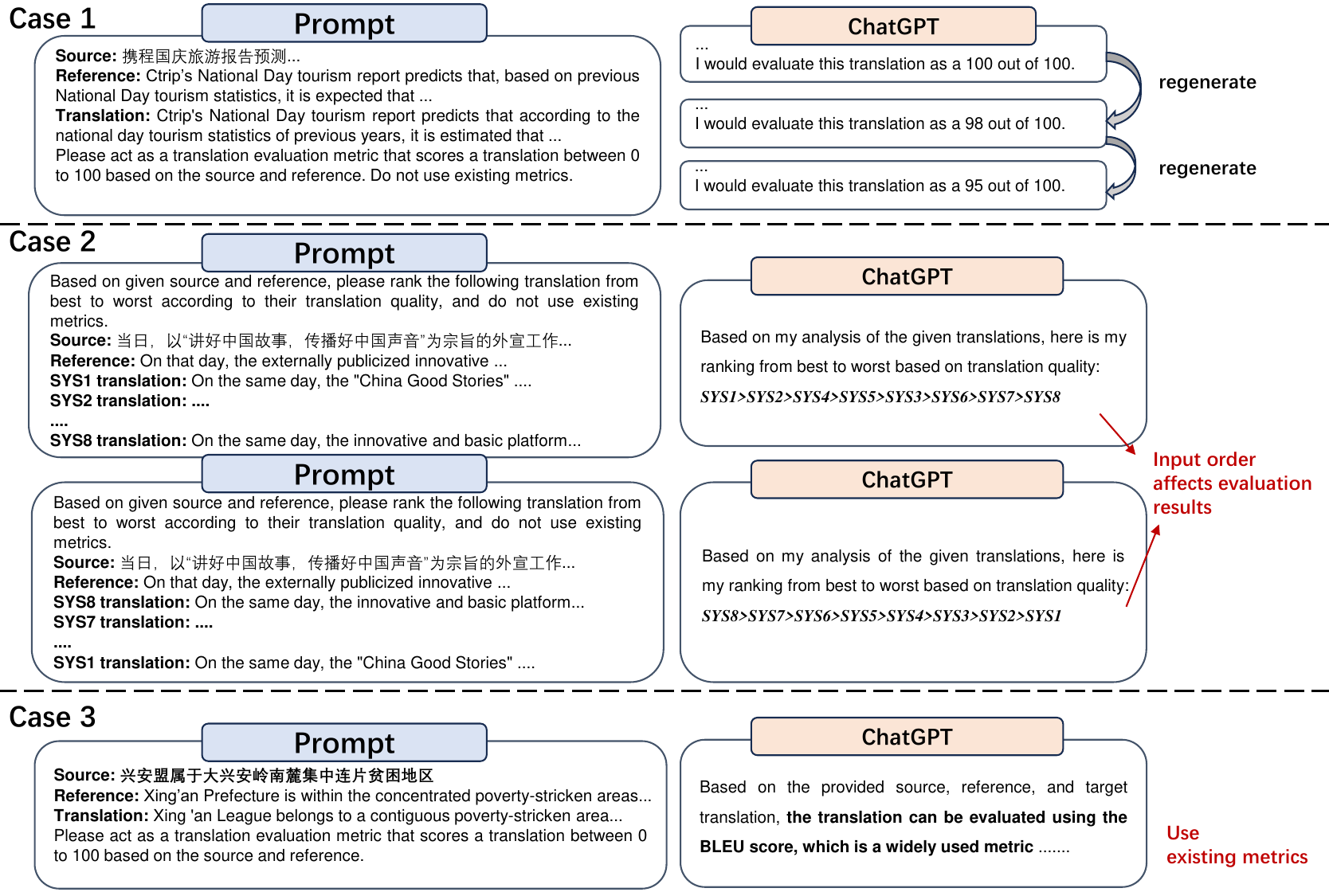}
\centering
\caption{\textbf{Case study of potential issues in LLMs}. All three cases are from GPT-3.5-Turbo model ("ChatGPT").  \textbf{Top}: LLM exhibits variations in its responses upon multiple regenerations; \textbf{Medium}: different input order of samples may affect the judgment of LLM; \textbf{Bottom}: LLM sometimes relies on existing metrics during translation evaluation.}
\label{fig:casestudy}
\end{figure*}

In Figure~\ref{fig:casestudy}, we list several typical issues with the case study that should be aware of when using LLMs such as ChatGPT as translation evaluators.

\subsection{Potential instability in the responses without temperature control}

\hl{\textbf{Issue:}}
When evaluating translations using LLMs, the generated responses may vary significantly. See in \textbf{Case 1}, we regenerate several responses with the same input and obtain 3 different scores (98, 95, 100) for the translation.

\noindent\hl{\textbf{Solution:}}
We control the temperature parameter to mitigate the variability in LLM judgments. Accordingly, for all experiments detailed in this paper, we set the temperature to 0 for \textbf{GPT-3.5-Turbo}. For the other two models, namely \textbf{Llama2-70b-Chat} and \textbf{Mixtral-8x7b-Instruct}, we opted for a temperature setting of 0.05 since the inference parameter from these two models should be above zero.

\subsection{Input order bias when evaluating multiple translations simultaneously} \label{sec:inputbias}

\hl{\textbf{Issue:}}
An alternative prompting strategy is to present multiple translations together as a single input to LLMs for evaluation, reducing the number of queries and potentially saving budget. However, we observe a bias where
translations presented earlier tend to get higher scores compared to those presented later. As shown in \textbf{Case 2}, we provide 8 translations along with their corresponding source and reference sentences. At the first time, we present the translations sequentially and ask LLM to rank them according to their translation quality. Then, we reverse the order of translations and obtain an entirely different sequence of ranks. 

\noindent\hl{\textbf{Solution:}}
The contradictory results may be attributed to the auto-regressive nature of the decoder model, which gives more attention to the latter input, potentially leading to greater identification of errors for the translation input later. Therefore, we recommend that researchers input one translation at a time instead of providing multiple translations. 

\subsection{LLMs may generate invalid answers for all prompting strategies} \label{sec:existmetric}

\hl{\textbf{Issue:}}
We observe that in certain cases, LLMs may not function as translation evaluators that may produce invalid answers with textual explanations. A typical case is illustrated in \textbf{Case 3}, where ChatGPT tends to prioritize the BLEU score instead of offering judgments based on its inherent capabilities.

\noindent\hl{\textbf{Solution:}}
We follow the method mentioned in \citet{kocmi-federmann-2023-large} for handling invalid answers, where we introduce randomness to LLMs by iteratively increasing the temperature. Subsequently, we take the first response that falls within the expected score range.

\end{document}